\documentclass[letterpaper,journal]{IEEEtran}
\usepackage{amsmath,amsfonts}
\usepackage{algorithmic}
\usepackage{algorithm}
\usepackage{array}
\usepackage[caption=false,font=normalsize,labelfont=sf,textfont=sf]{subfig}
\usepackage{textcomp}
\usepackage{stfloats}
\usepackage{url}
\usepackage{verbatim}
\usepackage{graphicx}
\usepackage{cite}
\usepackage{multirow}

\DeclareMathOperator*{\argmax}{arg\,max}

\begin{document}

\title{Lifelong Learning for Fog Load Balancing: A Transfer Learning Approach}


\author{Maad~Ebrahim,
        Abdelhakim~Senhaji~Hafid,~\IEEEmembership{Member,~IEEE,}
        and~Mohamed~Riduan~Abid,~\IEEEmembership{Member,~IEEE}%
\thanks{M. Ebrahim and A. S. Hafid are with the NRL, Department of Computer Science and Operational Research, University of Montreal, Montreal, QC H3T-1J4, Canada (e-mail: maad.ebrahim@umontreal.ca; ahafid@iro.umontreal.ca).}%
\thanks{M. R. Abid is with TSYS School of Computer Science, Columbus State University, Columbus, GA 31907, USA}
\thanks{Corresponding author: Maad Ebrahim (maad.ebrahim@umontreal.ca).}}

\maketitle

\begin{abstract}
Fog computing emerged as a promising paradigm to address the challenges of processing and managing data generated by the Internet of Things (IoT). Load balancing (LB) plays a crucial role in Fog computing environments to optimize the overall system performance. It requires efficient resource allocation to improve resource utilization, minimize latency, and enhance the quality of service for end-users. In this work, we improve the performance of privacy-aware Reinforcement Learning (RL) agents that optimize the execution delay of IoT applications by minimizing the waiting delay. To maintain privacy, these agents optimize the waiting delay by minimizing the change in the number of queued requests in the whole system, i.e., without explicitly observing the actual number of requests that are queued in each Fog node nor observing the compute resource capabilities of those nodes. Besides improving the performance of these agents, we propose in this paper a lifelong learning framework for these agents, where lightweight inference models are used during deployment to minimize action delay and only retrained in case of significant environmental changes. To improve the performance, minimize the training cost, and adapt the agents to those changes, we explore the application of Transfer Learning (TL). TL transfers the knowledge acquired from a source domain and applies it to a target domain, enabling the reuse of learned policies and experiences. TL can be also used to pre-train the agent in simulation before fine-tuning it in the real environment; this significantly reduces failure probability compared to learning from scratch in the real environment. To our knowledge, there are no existing efforts in the literature that use TL to address lifelong learning for RL-based Fog LB; this is one of the main obstacles in deploying RL LB solutions in Fog systems.
\end{abstract}


\section{Introduction}
\label{sec:intro}
The surge in demand for swift data processing and real-time responses due to the expansion of Internet of Things (IoT) devices and applications is unprecedented. IoT systems generate enormous data volumes, which require immediate analysis and decision-making capabilities. However, conventional cloud-based solutions struggle to cope with the sheer data volume and low-latency demands of IoT applications. In response to these challenges, Fog computing has emerged as a promising approach by moving computing resources closer to the network edge. Fog Load Balancing (LB) is a crucial element of Fog Computing, which efficiently allocates resources and enhances the overall efficiency and scalability of IoT applications.

LB can be performed by offloading workloads from heavily loaded nodes to neighboring nodes with more computing capacity, hence better serving the users with efficient utilization of Fog Computing capacity. This helps address the uneven distribution of application workloads between the different Fog nodes within the Fog network, which in turn reduces the overall system response time. Nonetheless, LB must efficiently perform while maintaining the privacy of Fog nodes, which is often required by Fog service providers. To avoid the complexity of privacy solutions \cite{ebrahim2022blockchain}, LB algorithms must work in partially observable environments, rendering the task more challenging.

\begin{figure}[!htbp]
\centering
\includegraphics[trim=280 50 0 40, clip, width=.48\textwidth]{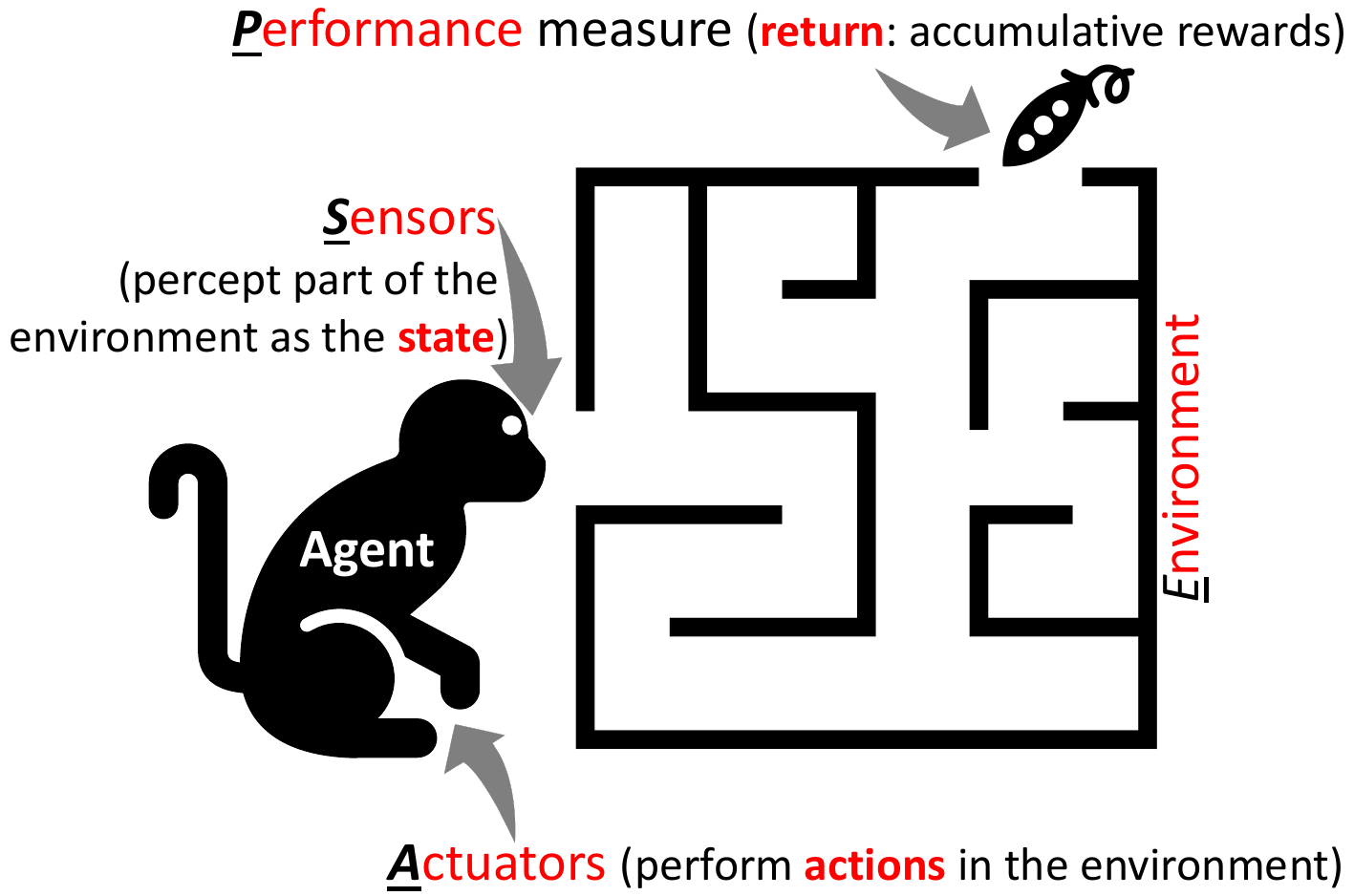}
\caption{RL in the PEAS Framework.}
\label{fig:PEAS_1}
\end{figure}

Intelligent LB solutions are a crucial part of Edge Intelligence \cite{ABID}, and can be achieved through intelligent rational agents through the PEAS (Performance measure, Environment, Actuators, and Sensors) framework. Reinforcement Learning (RL) agents, for example, can be viewed within the PEAS framework as agents that seek to maximize/minimize a performance measure, i.e., the return, by taking actions in their environments through actuators while using sensory input to make their decisions (see Fig. \ref{fig:PEAS_1}). Many intelligent agents have a learning component that allows them to adapt and improve over time. They update the agent's knowledge and decision-making strategies based on feedback from the environment through the performance measure.

Compared to tabular RL, Deep RL (DRL) addresses privacy concerns and partial observability by allowing agents to make informed decisions based on limited information, safeguarding sensitive data while achieving effective performance. Deep Neural Networks (DNN) perform function approximation in DRL algorithms to approximate policies and/or value functions, enabling agents to learn complex mappings from the limited information to actions without explicitly knowing the complete state of the environment. In addition to privacy awareness, DRL-based solutions can quickly adapt to minor traffic changes in real-world settings, unlike standard rule-based or optimization-based methods.

A major advantage of DRL methods compared to iterative or heuristic solutions is that once the agent has undergone pre-training, it becomes adept at providing real-time recommendations via simple arithmetic or array operations for DNN inference. Conversely, the training phase of a DRL system is more intricate compared to conventional techniques. Hence, in real-life implementations, it makes sense to stop training when the agent converges to the optimal solution since it can not get any better. This can save the huge cost of continuous training, which is not practical in real-world applications. Instead of adapting the agent to significant environmental changes through continuous training, the agent can detect these changes and adapt through fine-tuning when needed.

To learn and converge to a reliable control policy, a large amount of training data (interactions/experiences) is required, and hence longer training time. For instance, solving Atari games with DRL agents requires millions of interactions with the environment. This makes training and retraining very expensive and may even make such solutions impractical for complex real-world applications. Fortunately, Transfer Learning (TL) can help speed up the training and retraining processes by transferring useful information that was learned in source tasks to assist the learning process in target tasks. It improves data efficiency, speeds up learning, and provides better generalization compared to learning from scratch. 

TL can be used to initialize agents in new Fog networks by leveraging the expertise of pre-trained agents from existing Fog networks. This process involves transferring information from expert agents to new agents, jump-starting their learning process, and accelerating their performance. Information transfer encompasses various aspects, such as model weights and experiences. Notably, the learning process, including training and fine-tuning, often occurs in resource-rich devices, where computational capabilities are abundant. Once the agents are trained, inference agents can be deployed in devices with limited resources, such as IoT devices and smartphones. This strategy optimizes the efficiency of Fog networks, allowing for the efficient execution of real-time actions with a high level of performance and adaptability.

\begin{figure*}[!htbp]
\centering
\includegraphics[trim=30 360 30 0, clip, width=\textwidth]{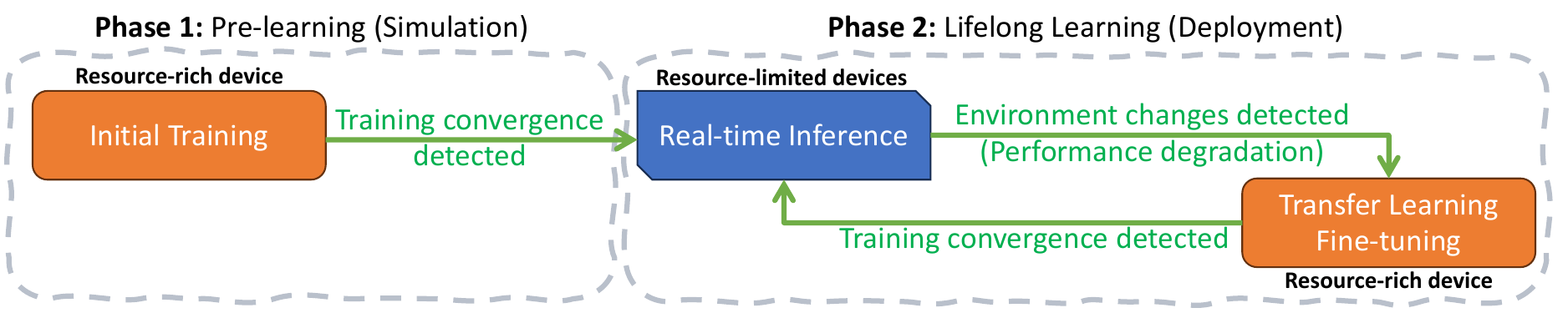}
\caption{Lifelong learning for RL solutions using TL.}
\label{fig:lifelong}
\end{figure*}

Hence, TL can be used to provide efficient lifelong learning for Fog LB agents, where agents experience sequences of tasks with different levels of difficulty. Instead of lifelong training, i.e., continuous training even after convergence, lifelong learning is composed of sequences of training and inference epochs (see Fig. \ref{fig:lifelong}). Lightweight inference agents provide real-time inference during deployment, and only retrained in case of significant environmental changes. Instead of retraining from scratch, the agents are fine-tuned using TL from their previous level of expertise speeding up the learning process. In addition, TL can help initialize the agents in simplified environments, simulation environments, or from agents already deployed in similar environments, instead of initializing them randomly from scratch. Then, these agents can be fine-tuned in their target real-world environments, which helps reduce possible failures during training with serious consequences.

To the best of our knowledge, lifelong learning of RL agents to perform LB in Fog environments has not been studied before. In addition, the impact of various TL techniques on improving LB in Fog environments has never been investigated in the literature. To fill this gap, we study in this paper the applicability of TL techniques to enhance the performance of Privacy-Aware RL (PARL) solutions for Fog LB \cite{ebrahim2023privacyaware}. Three different flavors of TL are studied, i.e., model weights transfer (knowledge), replay buffer transfer (experience), and full agent transfer (both knowledge and experience). These approaches are compared against retraining from scratch and against using the first trained agent without retraining it. We study the adaptation of these approaches to increases in task difficulty over time, i.e., significant increases in workload generation rates, while enhancing training time, performance, and consistency. 

To summarize, this paper addresses the challenges that arise from deploying RL agents in real-world Fog networks, including action delay, dynamic adaptation to significant environment changes, and privacy. These challenges are addressed through the key contributions of this paper, which can be summarized as follows:
\begin{itemize}
    \item Introducing a framework for lifelong learning of RL LB agents within Fog Computing environments. This framework facilitates real-time decision-making through the utilization of lightweight inference models.
    \item Investigating the efficacy of various TL techniques to enhance the adaptability of RL agents to substantial environmental alterations. This enhancement results in reduced training time and improved performance when retraining agents to cope with such changes.
    \item Enhancing the training efficiency and performance of RL agents in partially observable environments, where the agents are required to provide efficient load balancing while maintaining the privacy of Fog load and resource information.
\end{itemize}

The rest of the paper is organized as follows. Section \ref{sec:back} presents the background and the related work. Section \ref{sec:algo} presents the system model and the proposed approach. Section \ref{sec:eval} presents the performance evaluation. Finally, Section \ref{sec:conclusion} concludes the paper.

\section{Background}
\label{sec:back}

\subsection{Fog Load Balancing}
\label{subsec:fogLB}
Fog computing is a decentralized computing paradigm that extends cloud computing capabilities to the edge of the network, closer to the data sources and end-users. It aims to overcome the limitations of cloud-centric architectures by distributing resources, services, and applications across a distributed computing network that includes cloud data centers, edge devices, and intermediate Fog nodes. Fog computing enables efficient data processing, real-time analytics, and low-latency services; it makes it ideal for latency-sensitive and bandwidth-intensive applications, such as IoT, augmented reality, and smart city deployments.

LB in Fog Computing refers to the dynamic allocation and distribution of computational tasks and network traffic across Fog nodes; the objective is to achieve efficient resource utilization, minimize response time, and enhance the quality of service (QoS) for end-users. LB algorithms aim to evenly distribute the workload among Fog nodes, avoiding overloading or underutilization of resources. Efficient LB is critical in Fog Computing environments; it helps prevent bottlenecks, improve system performance, and ensure reliable and responsive services for IoT devices and end-users.

The benefits of efficient Fog LB to IoT applications include:
\begin{itemize}
    \item \textbf{Improved Performance}: LB algorithms optimize resource allocation, ensuring that computational tasks are evenly distributed across Fog nodes. This results in faster response times, reduced latency, and improved overall performance of IoT applications.
    \item \textbf{Enhanced Scalability}: By effectively distributing workloads, Fog LB enables the system to scale seamlessly with the increasing number of IoT devices and data volumes. Scalability is vital in accommodating the growth of the IoT ecosystem and maintaining high-quality user experiences.
    \item \textbf{Load Optimization}: LB algorithms analyze the workload distribution and adjust the assignment of tasks dynamically. This optimization prevents resource underutilization or overutilization, improving the overall efficiency of Fog Computing environments.
    \item \textbf{Fault Tolerance}: Fog LB algorithms can handle failures or performance degradation of individual Fog nodes. By redirecting tasks to alternative nodes, LB ensures fault tolerance and high availability, reducing the impact of node failures on IoT applications.
\end{itemize}

However, LB in Fog Computing faces several challenges; they include the heterogeneity of Fog nodes with varying computational capacities, dynamic changes in the availability of Fog resources, heterogeneity of workloads with varying computational requirements, and limited network bandwidth. Traditional LB techniques, such as round-robin or random assignment, may not be sufficient to handle the complexities of Fog Computing environments. Therefore, there is a need for intelligent self-learning LB algorithms that can optimize resource allocation, adapt to changing conditions, and provide efficient and reliable services.

\subsection{Reinforcement Learning}
\label{subsec:rl}
RL is a branch of machine learning that focuses on training intelligent agents to make sequential decisions in an environment to maximize cumulative rewards. RL algorithms learn through interaction with the environment, receiving feedback in the form of rewards or penalties for their actions. The agent learns to identify the optimal policy or action-selection strategy to maximize its long-term rewards. RL has shown great success in various domains, including game playing, robotics, and resource allocation problems.

DRL agents can work in partially observable dynamic environments \cite{pomdps}, where they learn from their own experiences to continually adapt to changing conditions. They might find optimal/near-optimal solutions without observing all the required information from the environment; this makes them effective in situations with complex or unknown system information, like providing efficient LB while maintaining the privacy of Fog information. In addition, DRL substitutes tabular RL with DNN function approximation, enabling it to work on problems with large state spaces and sparse reward functions. 

By leveraging the generalization capabilities of DNN, DRL can effectively handle unseen states, since most encountered states are likely to be novel. Although RL models require significant time and resources for training, once trained, lightweight versions of the models can be extracted and deployed with minimal compute and storage requirements. However, these models will still have to be retrained to cope with significant dynamic environment changes. Sequences of inference and fine-tuning epochs for RL agents allow for efficient lifelong learning in real-world deployment. This helps save a significant amount of resources compared to continuous training in real-world deployments. In such cases, TL becomes crucial as it allows RL agents to leverage previous knowledge to accelerate adaptation to unfamiliar settings, such as changes in Fog network size, available Fog resources, application requirements, and demands \cite{Harmonizing}.

\subsection{Transfer Learning}
\label{subsec:tl}
TL is a machine learning technique that involves using the knowledge learned on a specific task as a starting point for a different but related task \cite{ebrahim2019transfer}. It is commonly used in natural language processing and computer vision. For example, a pre-trained model that was trained on a large dataset of images could be used as a starting point for a new model to classify a new set of images with similar features. Hence, TL can save time and resources as it can speed up training time and improve performance compared to learning from scratch.

\begin{figure}[htbp]
\centering
\includegraphics[width=0.48\textwidth]{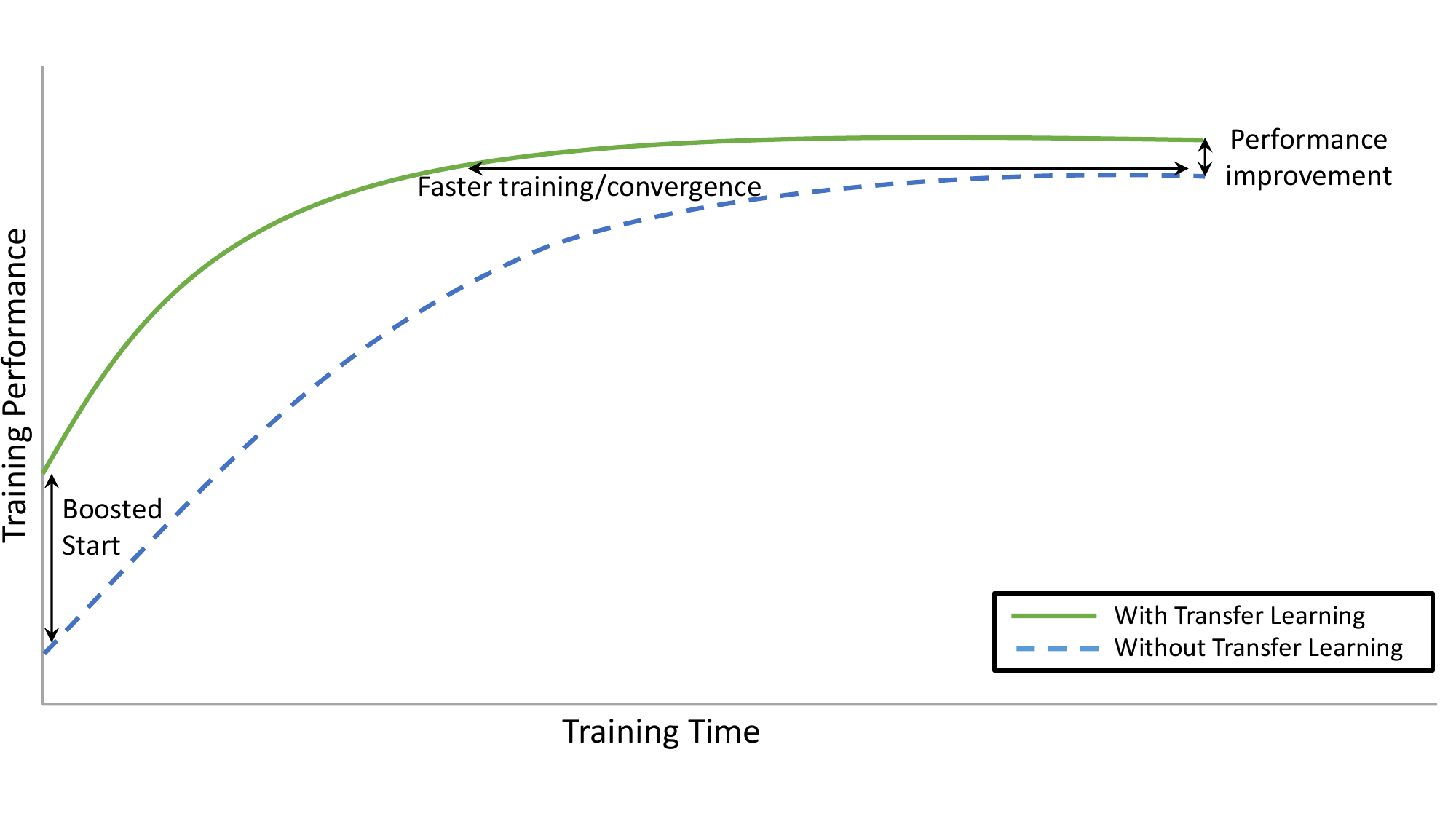}
\caption{The benefits of TL for RL agents.}
\label{fig:tl_benifits}
\end{figure}

In RL, TL involves using the knowledge of a pre-trained agent to improve the learning speed and performance of a new agent for a similar task or environment. It enables the agent to adapt its knowledge to the target domain by leveraging the shared features or representations from the source domain. Fig. \ref{fig:tl_benifits} shows the benefits of successful TL compared to starting from scratch, i.e., boosted initial performance, faster convergence, and better ultimate performance. In addition, here are some other benefits of TL to RL agents:
\begin{itemize}
    \item \textbf{Speeding up Learning:} Fine-tuning a pre-trained RL agent using TL can significantly accelerate the learning process in the new environment. Instead of starting from scratch, the agent can leverage its existing knowledge and build upon it, reducing the time required to learn effective policies. This is especially important when the changes in the environment are significant, as the agent can quickly adjust its behavior based on the previously learned information.
    \item \textbf{Knowledge Retention} TL allows the RL agent to retain and transfer valuable knowledge from the previous environment to the new one. By leveraging the shared features or representations learned in the source environment, the agent can avoid relearning basic skills or low-level behaviors. Thus, it can focus on adapting its policies to the specific characteristics of the new environment, leading to faster convergence and more efficient learning.
    \item \textbf{Sample Efficiency:} Fine-tuning an RL agent using TL can significantly reduce the sample complexity required in the new environment. The agent can start with a pre-trained model that has already learned useful representations, allowing it to quickly adapt its policy using a smaller number of interactions with the new environment. This is especially beneficial when gathering training data is challenging or expensive, as it can save valuable time and resources.
    \item \textbf{Robustness and Generalization:} TL improves the robustness and generalization capabilities of RL agents in the face of significant environmental changes. By fine-tuning the agent with TL, it can adapt its policies to variations, noise, or uncertainties present in the new environment. The agent can leverage its prior knowledge to generalize its behaviors effectively, even in situations that were not encountered during the initial training.
    \item \textbf{Incremental Learning:} In scenarios where the environment undergoes gradual changes over time, TL facilitates incremental learning for RL agents. Instead of retraining the agent from scratch every time the environment changes, TL allows for an incremental update of the agent's policy based on the new environment. This allows for lifelong learning and adaptation, making the RL agent more responsive to evolving conditions.
\end{itemize}

TL can show improvements on target tasks with more challenging system dynamics than their related source tasks, i.e., the target task becomes harder to solve in an incremental way. An example of a target task with increased difficulty is the well-known mountain car game in RL literature \cite{singh1996reinforcement}, where the agent has to move a car along a curve to reach the top of the curve using three actions, i.e., Forward, Neutral, and Backward. To increase the difficulty of the game, we can change the power of the motor, the surface friction, the starting position, or even let the agent control two cars simultaneously. Because real-world environments are constantly changing, TL plays a crucial role in developing realistic RL solutions. It is particularly valuable with significant environmental changes, as it enables the fine-tuning of RL agents for quick and efficient adaptation.

It is important to identify the type of information to be transferred between source and target tasks. Different types of information can be better or worse depending on task similarity. For example, low-level information can be transferred across more similar tasks, while high-level information might work better across less similar tasks \cite{taylor09}. The transfer of high-level information can include the transfer of general rules or advice inspired by the agent that was trained in the source task. On the other hand, low-level information includes replay buffer experiences, the action-value function, the policy, or a combination of them. 

TL via Policy Reuse was proposed in \cite{PolicyReuse}, where policies from source tasks are directly reused to build the target policy. With DRL the knowledge learned by the agent is represented by the neural network weights, which can be transferred from the source agent's policy to initialize the target agent's policy. In addition to policy reuse, the experience of the source agent, represented by its replay buffer, can be transferred to initialize or complement the replay buffer of the target agent. 

It is important to carefully choose what information to transfer to avoid TL negative effects \cite{Harmonizing, Survey}, also called negative transfer or negative guidance. Negative transfer depends on the differences between source and target tasks and occurs when the information from source tasks hinders or interferes with learning target tasks.

\begin{figure*}[htbp]
\centering
\includegraphics[width=0.9\textwidth]{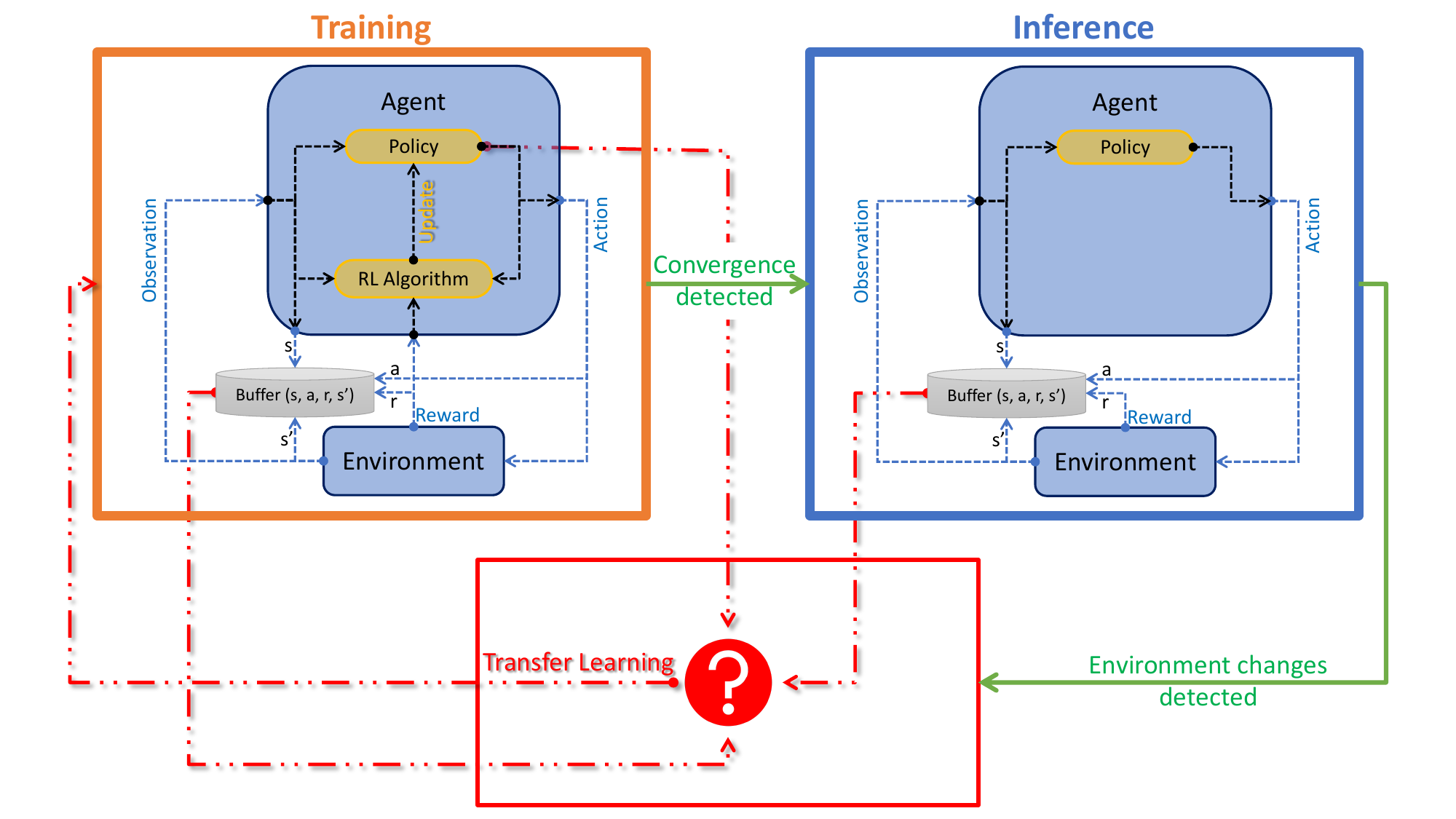}
\caption{Knowledge and/or experience TL for RL agents.}
\label{fig:tl}
\end{figure*}

Figure \ref{fig:tl} shows the need for TL for RL-based realistic solutions, where training and inference MDPs significantly differ over time. During the expensive training phase, the agent updates its policy using the RL algorithm until convergence. However, during deployment, the learned policy in lightweight inference agents can be used even on resource-limited devices since no training is needed. In case of significant changes in the environment, the agent adapts by fine-tuning its model to adapt to these changes. Besides the experience that is gathered during deployment, TL techniques can include previous experiences and/or previous knowledge. Hence, TL can reduce the training time to quickly adapt to these changes, improve the performance of the agent, retain previous knowledge for better robustness and generalization, and enhance the ability to perform incremental lifelong learning.

\subsection{Related Work}
\label{subsec:related}
The idea of TL has, only recently, been applied to RL tasks. This motivated researchers to study the differences between TL techniques in the RL domain. Taylor et al. \cite{taylor09}, for example, classified TL methods in terms of their capabilities and goals. Zhu et al. \cite{deepTL} investigated the recent progress of TL approaches focusing more on DRL rather than tabular RL. Based on the required number of training interactions, Zhu et al. \cite{deepTL} categorized TL approaches into the following classes:
\begin{itemize}
    \item \textbf{Zero-shot transfer:} The agent is directly applicable to the target task without further training interactions.
    \item \textbf{Few-shot transfer:} The agent requires a few interactions with the target environment to completely adapt to the target task.
    \item \textbf{Sample-efficient transfer:} The agent needs to interact with the target environment long enough to converge. However, it still benefits from TL to become sample efficient faster than starting from scratch.
\end{itemize}

Sample efficiency by acquiring sufficient interactions is very important for RL agents; however, it can be prohibitive in life-threatening domains such as self-driving vehicles and healthcare. TL provides sample efficiency in such situations by pre-training, for example, RL agents of self-driving vehicles in simulated environments before fine-tuning them in real-world environments \cite{finn2019meta}. In addition, TL helps when the reward signal is sparse or delayed in the target environment \cite{deepTL}. In such cases, the agent is first pre-trained in a simplified environment before being fine-tuned in the actual environment with sparse rewards.

The idea of lifelong learning was first proposed in 1996 \cite{easier}, where learners experience streams of learning tasks. Sutton et al. \cite{suttonTracking} expanded this idea to RL problems; they stated that agents are actually performing sequences of tasks when interacting with the real world for long periods. Lifelong learning is not necessarily done through continuous training, where agents continue training to adapt to environmental changes, even after convergence. Instead, lightweight inference agents can be deployed and only fine-tuned in case of significant changes that can degrade their performance. The fine-tuning process is done using TL from the previously trained agents allowing for faster and less demanding training.

Selfridge et al. \cite{SelfridgeRobotics} investigated the adaptability of RL agents to environment changes while benefiting from progressive training sequences. Those sequences start with easier versions of the tasks that gradually get harder. They found that it was faster to learn to balance a pole on a cart by gradually making the transition function of the task harder over time. First, the learner was trained on a long and light pole, and once learned to balance it successfully, the task was made harder by shortening and making the pole heavier. The total time spent training those sequences, while reusing the learned knowledge, was faster than training on the hardest task directly. 

Similarly, Asada et al. \cite{vision} devised a maze that presents a gradual increase in difficulty. But, rather than modifying the dynamics of the maze, they incrementally relocated the starting point, i.e., the initial state, further and further away from the endpoint. Using TL between those sequences enabled the agent to navigate toward the exit faster and more efficiently compared to learning to traverse the entire maze at once. 

TL is also used for control problems, such as LB problems. For example, Wu et al. \cite{DataEfficient} used TL to create data-efficient DRL agents that perform LB for communication networks. Instead of starting from scratch, they boosted the performance of new base station agents, with a limited amount of training data, using TL from a policy bank that is collected from previously trained base station agents. Their results did show that TL significantly improved the system performance and made RL agents more robust to environmental changes.

Mechalikh et al. \cite{Fuzzy} used TL when a new orchestrator joins existing orchestrators in a given cluster of Edge devices to adapt to mobility and power consumption changes. The new orchestrator merges the knowledge of nearby orchestrators to initialize itself. Compared to random initialization, TL increased the success rate, i.e., minimized failure, especially immediately after the selection of the new orchestrator.

Yang et al. \cite{IoV} used TL to effectively enhance the learning efficiency and convergence speed of their RL resource management solution that supports delay-sensitive Internet-of-Vehicles (IoV) services. New vehicles are encouraged to utilize the knowledge of neighboring expert vehicles with similar characteristics. Learning vehicles add the actions of expert vehicles to their actions before greedily selecting an action from that combined set of actions. TL rate decreases over time, to allow the learning vehicle to gradually become independent from the expert vehicle.

Sun et al. \cite{Green} integrated TL into their DRL resource management solution in Fog Radio Access Networks (RANs). Instead of training from scratch, they transferred the weights between agents that were trained in similar environments. Their results showed that TL helps achieve performance similar to training from scratch but with much less training time. They also showed that TL can lead to negative guidance on target tasks when the similarity between the source and target tasks is low, i.e., the discrepancy in nodes' resources between source and target tasks is amplified.

We conclude that TL is a promising approach to enhancing the efficiency and adaptability of RL agents in a variety of domains. However, there is a noticeable gap in the existing literature when it comes to applying TL to Fog load balancing problems, especially for the purpose of providing quick recovery from performance degradation due to significant changes in the environment. In this paper, we address this gap by presenting a novel lifelong learning approach for RL agents through TL in Fog computing environments. 

As seen in prior research for LB problems \cite{Green, IoV, Fuzzy, DataEfficient}, the use of TL is limited to merely transferring information from expert agents to newly added agents. Instead, we use TL in our lifelong learning framework to continuously enhance the training performance of the agent throughout its lifetime, leveraging its previous experiences and learned policies. Additionally, we propose the transfer of simulation experiences and acquired knowledge from simulated environments to real-world scenarios, further enriching the agent's adaptability and improving its performance while minimizing failure probability during initial training.

\section{The TL-PARL Approach}
\label{sec:algo}

\subsection{The Fog System Model}
\label{sec:model}
A Fog system has a set of $N = n_1, n_2, \cdots, n_z$ nodes; each node $n_x$ is defined by its compute $IPT_x$ and memory $RAM_x$ resources. Nodes are connected through $L = l_1, l_2, \cdots, l_z$ bidirectional links; each link $l_x$ is characterized by the pair ($n_i$, $n_j$) that it connects, its bandwidth $BW_x$, and its propagation delay $PR_x$. 

A set of distributed applications $DA = da_1, da_2, \cdots, da_z$ simultaneously run in the system; each distributed application $da_x$ is represented by a set of modules $M_x$ and a set of dependencies $D_x$ between these modules, i.e., called messages, workloads, jobs, or requests. Generally, messages flow in loops, where the most common is the immediate Fog feedback loop for every source message. The cloud loop is also common to provide intermittent feedback on aggregated source data.

Hence, a Fog system is defined by the set of nodes $N$, links $L$, and distributed applications $DA$ running in the system. Fog nodes have heterogeneous resources and can directly communicate with each other at the Fog level without the need to go through the Cloud. IoT devices that are connected to the same communication link can be treated as a single source $c$ since a large number of IoT devices are often geographically deployed together. Undoubtedly, conceptualizing large numbers of IoT devices as smaller numbers of IoT clusters $|C|$ is important for implementing real-time LB solutions in such large and complex environments.

Similarly, instead of dealing with an infinite number of workload compute requirements, it is more efficient to work with smaller numbers of workload categories $|W|$, for example, light, medium, and heavy workloads. The agent can learn to perform the same action for all workloads from the same category $w$. In addition to simplicity, using discrete sets of source clusters and workload categories allows for dynamic adaptation to minor changes in IoT physical distribution and workloads' requirements, respectively.

Workloads are generated as a Poisson Point Process, and their generation rate is defined using an exponential distribution with a scale parameter $\beta$ \cite{MCDM}. The smaller the scale parameter, the more frequently these workloads are generated. Hence, the value of $\beta$ is decreased to model a significant increase in generation rates, i.e., increased difficulty for the Fog LB agent.

Nodes and network links are modeled with M/M/1 queuing models to capture the generated workloads \cite{YAFS}. The total number of jobs queued in the system is $\mathbb{Q} = \sum_{i=0}^N \mathtt{q}_i$, i.e., the summation of the number of jobs in each individual Fog node. To minimize the overall execution delay of the system, it is enough to minimize the time spent on these queues, i.e., waiting delay, which can be decreased through minimizing the number of jobs queued in Fog nodes \cite{ebrahim2023privacyaware}. 

\begin{figure*}[!htbp]
\centering
\includegraphics[trim=0 30 0 0, clip, width=\textwidth]{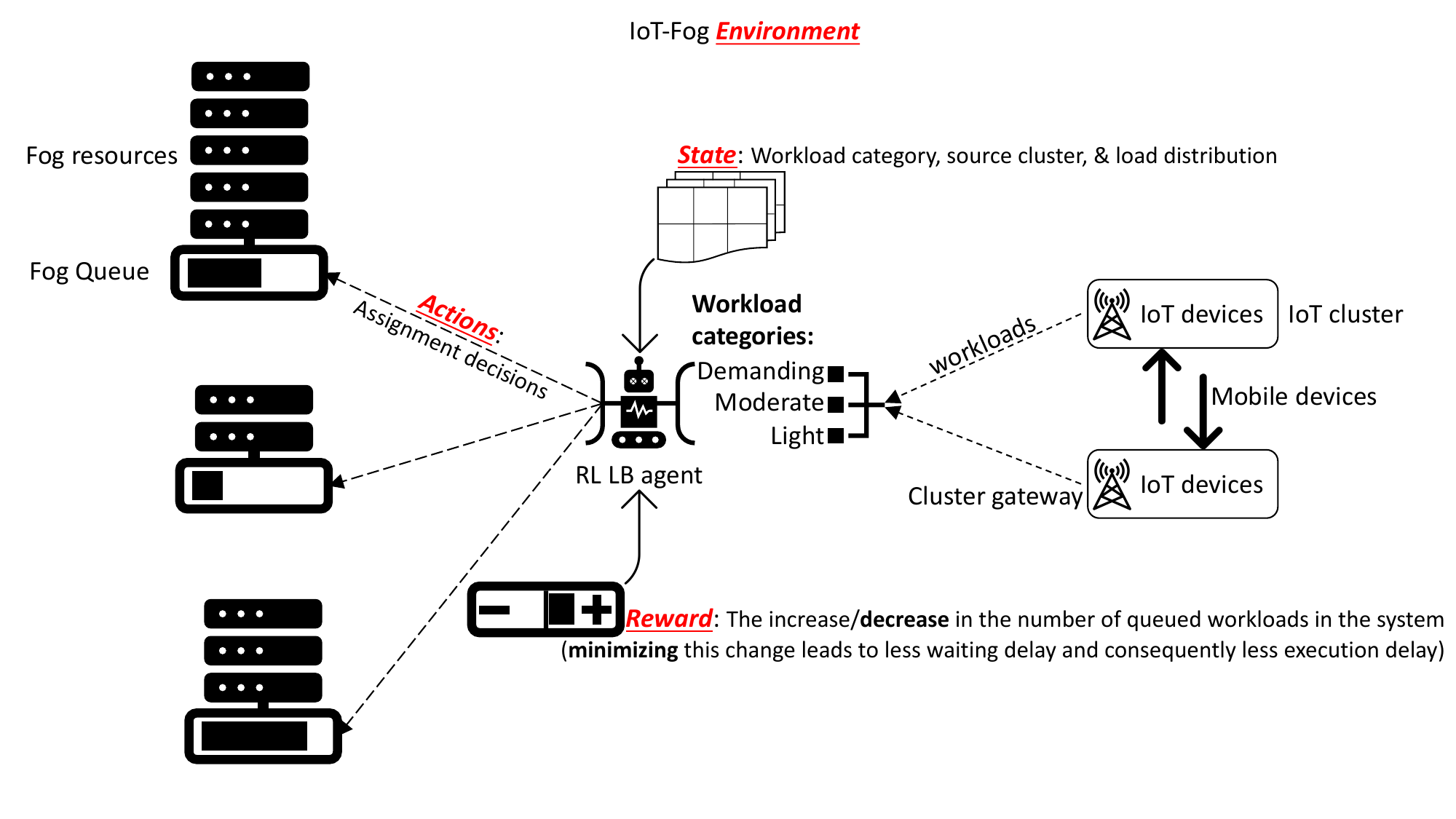}
\caption{PARL agent for Fog LB in the PEAS Framework.}
\label{fig:PEAS_2}
\end{figure*}

Hence, for every generated workload, the LB agent learns to select the best Fog node from the set of available Fog nodes $|N|$, i.e., from a total number of $|A|$ actions. These decisions are based on the workload category and the source cluster it was generated from (see Fig. \ref{fig:PEAS_2}). Fig. \ref{fig:PEAS_2} shows the interactions of the PARL agent with the Fog environment; the agent perceives a part of the environment as the state and performs actions by assigning workloads to Fog nodes in a way that maximizes its performance measure. 

The agent views groups of IoT devices, that are connected through the same gateway, as source clusters. This helps the agent adapt to changes in the physical distribution of IoT devices; these changes are viewed as changes in generation rate. In addition, the agent categorizes workloads into a discrete set of categories based on their requirements; this allows the agent to handle infinite variations in these requirements. The assignment decisions of the agent should maintain equal waiting times in different Fog nodes despite variations in their capabilities. Indeed, the agent receives the change in the number of queued workloads in the system as its reward and adds an internal representation of the distributed load to its state. For more details about the PARL agent and the Fog system model, the reader is referred to our previous contributions \cite{EBRAHIM2023resilience, ebrahim2023privacyaware}.

\subsection{The DDQL Algorithm}
\label{subsec:ddql}

To minimize the total execution time, the agent must assign workloads to Fog nodes in a way that minimizes the average waiting delay \cite{ebrahim2023privacyaware}. This is critical to avoid overflow of Fog heterogeneous resources and increase their resource utilization. Thus, we define the reward as the change in the number of queued jobs, i.e., $r = \Delta\mathbb{Q} = \mathbb{Q}_{t-1} - \mathbb{Q}_{t}$. Hence, the accumulative performance of the agent behavior, i.e., the expected return, is given by $G = \sum_{t=0}^{\infty} \gamma^t r_t$. This expectation evaluates the expected value of future states after following the current policy, where $\gamma$ defines the importance of future rewards compared to recent ones.

The agent needs to observe the load distribution of each workload category from every source cluster to every Fog node. Thus, we define the state by the source cluster, the workload category, and a local view of load distribution, i.e., $s = <c, w, d>$. $d$ is a 3D matrix with the following dimensions: number of clusters ($|C|$) $\times$ number of workload categories ($|W|$) $\times$ number of actions ($|A|$). To improve convergence and avoid numerical instability, $d$ is normalized to sum to 1, i.e., $\sum_{c=1}^{|C|}\sum_{w=1}^{|W|} \sum_{a=1}^{|A|} d = 1$.

Agents take actions according to the current environment state and their learned policy $\pi$. They train to learn the optimal policy $\pi^*$ by evaluating how good it is to take action $a$ in state $s$ and follow policy $\pi$ thereafter. This is called the Q-Value, Q-Function, or action-value function; it is defined for each state-action pair as $Q_\pi(s, a) = \mathbb{E}\left[G \mid s, a, \pi \right]$. The Q-function using the optimal policy, i.e., optimal Q-function $Q^*(s) = \max_\pi Q_\pi(s)$, cannot be less than the Q-function using any other policy, i.e., $Q_{\pi^*}(s) \geq Q_\pi(s)$.

DQL works in more complex, high-dimensional, and continuous state spaces by approximating the Q-values using DNNs instead of traditional Q-tables. With function approximation, a mapping of state-action pairs ($\phi$) is weighted by $\theta$ to calculate the $Q$-Function for each state-action pair, i.e., $Q(s, a) = \sum_{i=1}^d \theta_i \phi_i (s, a)$. This $Q$-Function is updated by minimizing a Temporal Difference (TD) error multiplied by a learning rate $\alpha$:
\begin{equation*}
    Q(s, a) \leftarrow Q(s, a) + \alpha \overbrace{\left[ \underbrace{r + \gamma \overbrace{\max_{a'}Q(s', a')}^{\text{Estimation of future value}}}_{\text{Temporal difference target}} - \enspace Q(s, a) \right]}^{\text{TD error}}
\end{equation*}

The TD error is computed by subtracting the $Q$-Function from a TD target; it is calculated by adding the immediate reward to the discounted estimation of the optimal future value. The best action in a given state is found by searching for the action that maximizes the $Q$-Function for that state, i.e., $\pi(a\mid s) = \argmax_aQ(s, a)$. Once the algorithm converges to the optimal $Q$-Function ($Q^*$), the optimal policy can be obtained as $\pi^*(s) = \argmax_aQ^*(s, a) ~ \forall s \in S$.

Multiple steps from a random history of samples, i.e., Experience Replay Buffer, are used to update the $Q$-Function instead of using the most recent step only; this avoids policy oscillation, instability, and divergence \cite{mnih2013playing}. To solve the common over-estimation problem in Q-Learning and DQL algorithms, Double-DQL (DDQL) uses a model $Q$ for action evaluation and a target model $Q'$ for action selection \cite{ddqn}. The $Q$ model is updated every training step, and periodically copied to $Q'$ every predefined number of steps, making the algorithm off-policy. Learning the optimal policy while following an exploratory policy for action selection helps balance exploration and exploitation.

\subsection{Privacy Awareness}
\label{subsec:privacy}

Hiding specific information about Fog nodes is a strategic decision for Fog service providers to enhance security, protect sensitive data, and maintain a competitive advantage in the Fog Computing market. By not disclosing everything about their Fog nodes, they can keep their edge over competitors. Conversely, revealing detailed information about the architecture, infrastructure, or software running on Fog nodes can make them more susceptible to targeted attacks. 

Hiding certain information about Fog nodes can be part of resource allocation strategies. For instance, users/applications prefer nodes with higher capacities and capabilities; decreasing the utilization of system resources. Therefore, it is best to optimally distribute tasks without revealing the specifics of each node. However, providers need to strike a balance between security and transparency to ensure users have enough information to make informed decisions about the services they receive. 

Defining the state without Fog resource/load information provides the required privacy for Fog service providers. It also makes the algorithm adaptable to minor environmental changes, i.e., changes in load demands and available resources. To maintain privacy while allowing the agent to make informed LB decisions, the agent keeps a local view of the distributed load. To recognize recently and frequently selected Fog nodes, the concept of vanishing normalization was proposed \cite{ebrahim2023privacyaware}. It gives higher values to recently selected Fog nodes for every given workload category and source cluster. 

Introducing a local view of load distribution with vanishing normalization allows the agent to capture the history of interactions in the environment without the need for more complex methods like recurrent neural networks \cite{deepRQL}. The load distribution is implemented as an array that is normalized to sum to 1 by dividing each element by the total sum of array elements. The normalization step takes place at each step after adding 1 to the array element representing the most recent decision, i.e., the assignment of the recently generated workload category from its source cluster. For more details, the reader is referred to Algorithm 1 in \cite{ebrahim2023privacyaware}.

Minimizing the waiting delay is the main factor in minimizing the overall system execution delay \cite{ebrahim2023privacyaware}. However, instead of providing the agent with the actual number of queued jobs in the state/reward representation as in Privacy-Lacking RL representations (PLRL) approaches, PARL reward representation uses the difference in the number of queued jobs between two consecutive decision steps. When trained long enough, the PARL state/reward representation achieves marginally superior performance to PLRL without explicitly using the actual Fog load information \cite{ebrahim2023privacyaware}. With higher workload generation rates, PARL agents outperformed PLRL agents with better and more consistent performance \cite{ebrahim2023privacyaware}.

Maintaining privacy makes the problem partially observable, which is harder to solve compared to fully observable RL due to the additional challenge of dealing with incomplete information about the environment. In Fully Observable RL, the agent has access to complete state information at each time step, allowing it to make informed decisions based on the current state. However, in partially observable RL, the agent only receives partial observations of the environment, making it difficult to accurately estimate the true state.

The added complexity of maintaining privacy requires the agent to spend more time before convergence, especially with harder tasks. PARL agents require more exploration to get insights into the dynamics of the environment without the need to fully observe all environmental information. This makes it more difficult to find the optimal trade-off between exploration and exploitation, which in turn delays convergence. TL can solve these problems and speed up the learning process by leveraging data from previously learned tasks. It reduces the required number of training interactions in target tasks given enough interactions in previous task sequences, even with partially observable environments.

\subsection{The TL Phase}
\label{subsec:tl_phase}
TL can be done using pre-trained agents as starting points for training new agents on related tasks. A pre-trained agent is used to initialize the $Q$-Network weights of a new agent before being fine-tuned on the new task. In addition, the interactions collected by the pre-trained agent, i.e., the experience trajectories in the replay buffer, are used to train the new agent on the new task. It is important to note that the source and target tasks must share a certain level of similarity to achieve an effective transfer of knowledge/experience.

To provide efficient lifelong sequential learning for Fog LB agents, TL can be incorporated between sequences of tasks with different levels of difficulty. The goal of TL here is to enhance performance with a smaller number of training interactions compared to learning from scratch. In this work, TL is used between three sequences of tasks with increased difficulty. The agent starts learning in an easy environment with a low generation rate. Then, the difficulty significantly increases to a medium generation rate before being increased again to high. To study what contributes the most to more efficient TL, we evaluate the following TL techniques:
\begin{enumerate}
    \item \textbf{No} TL: The agent trains from scratch after every significant change in generation rates.
    \item Training on the \textbf{First} generation rate only: The agent is only trained with the lowest generation rates, and used as it is with higher rates.
    \item Transferring \textbf{Buffer} Experience: Transfer previous experience to a new agent with randomly initialized weights. 
    \item Transferring \textbf{Weights} knowledge: The replay buffer is flushed before retraining the agent without altering the previously learned weights. 
    \item \textbf{Full} agent transfer: The agent continues training using its previous experience and knowledge, i.e., without flushing the buffer nor re-initializing its learned weights.
\end{enumerate}

\section{Performance Evaluation}
\label{sec:eval}

\subsection{Experimental Environment}
\label{subsec:env}
To evaluate our approach, realistic Fog topologies with unbalanced Fog resources and unbalanced workload distribution are simulated using a Discrete-event Simulator (DES), i.e., YAFS \cite{YAFS}. These topologies were created using a random graph generator that mimics the Internet Autonomous System \cite{networkx}, which is used to simulate interconnected flat Fog systems. Using betweenness centrality \cite{betweenness}, nodes are identified as Cloud, Fog, or IoT source clusters. To resemble unbalanced heterogeneous resources with unbalanced load distribution, the fewer compute resources a Fog node has the more IoT clusters it is connected to.

Three categories of workload compute requirements are used, i.e., heavy (resource-demanding), moderate, and light workloads. These workloads are generated as a Poisson Point Process using exponential distributions with three scale parameters, i.e., $\beta=100$, $\beta=150$, and $\beta=200$, to represent high, medium, and low generation rates, respectively. The proposed DDQL agent was implemented using TF-Agents \cite{TFAgents}; it is implemented to interactively work with the DES simulator to mimic practical deployment in real environments. 

\begin{figure}[htbp]
\centering
\includegraphics[trim=0 120 0 0, clip, width=.48\textwidth]{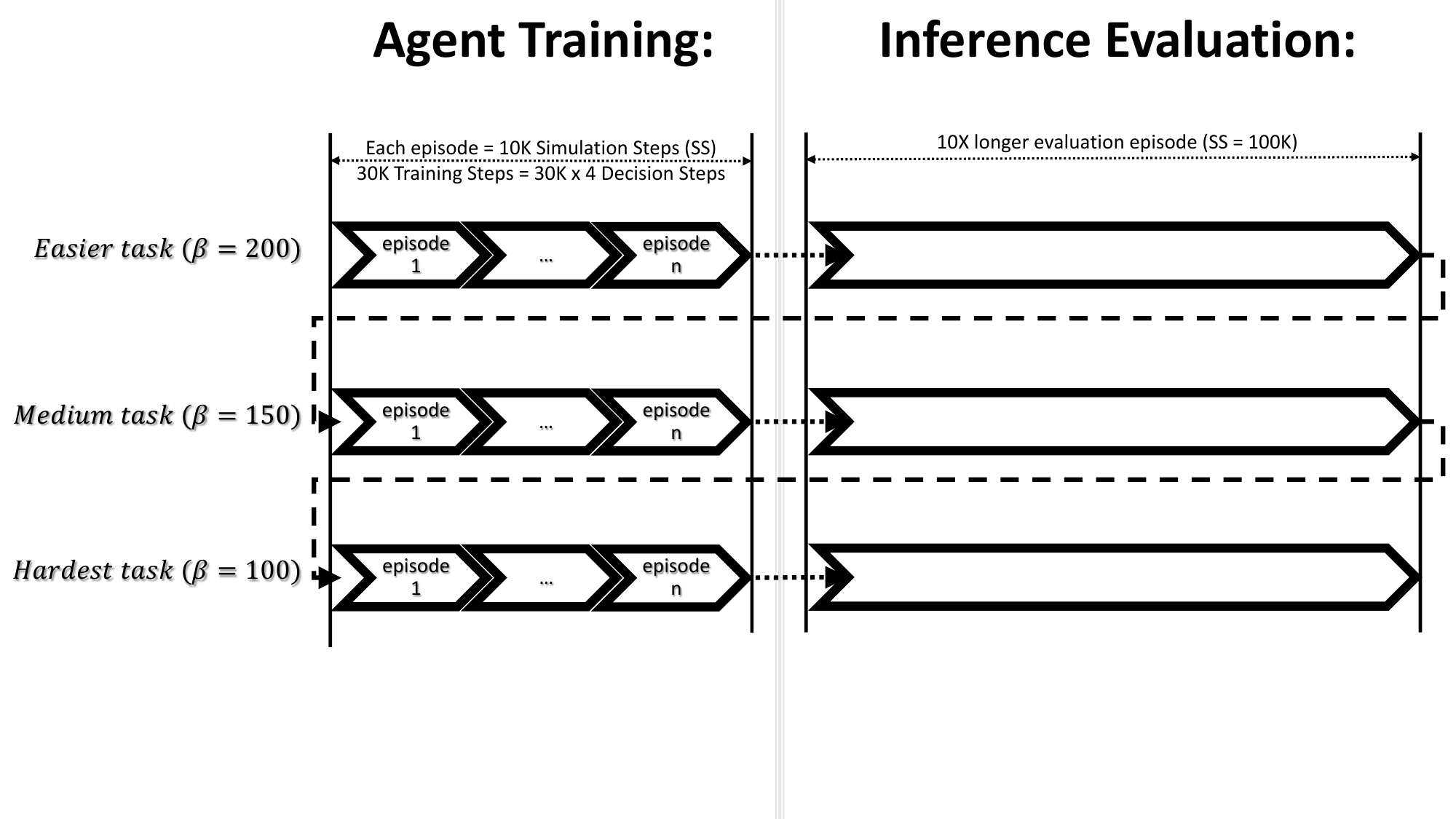}
\caption{Agent training and inference evaluation.}
\label{fig:workflow}
\end{figure}

To study which TL method can make PARL agents train and converge faster, agents are trained with 20\% of the number of training steps compared to our previous work \cite{ebrahim2023privacyaware}. Using $5\times$ less training was the minimum we could go for while still achieving the optimal solution using the PLRL agent. Therefore, instead of training the agents for 150K training steps, the agents in this work are trained for 30K training steps only. Agents start learning the easier task using a low generation rate (see Fig. \ref{fig:workflow}). Instead of using a single episode to train the agent, a series of episodes with 10K simulation steps each is used. This allows the agent to handle sparse and delayed rewards by focusing more on immediate outcomes. 

Since a load assignment decision in one step affects the outcomes of later steps, shorter episodes will help the agent perceive the temporal dependencies of these decisions. However, the agent's inference performance is evaluated using a single episode that is $10\times$ longer. After performing inference for 100K simulation steps, the generation rate instantly increases to medium, and then to high requiring the agent to adapt to these changes using one of the TL methods used in this work. 

The reader should differentiate between simulation steps, decision steps, and training steps, where simulation steps represent simulation time units corresponding to time units in real life. A decision step, on the other hand, occurs for each new job assignment decision, which can span over multiple simulation steps. Due to the random Poisson Process, the simulation time advances by a variable number of simulation steps between each two consecutive decision steps. Finally, a training step takes place every four decision steps, which is a common practice to allow the agent to collect and save a few experiences to the replay buffer before training on a random batch from that buffer. 

Simulations run on a DELL G5 Gaming laptop with an Intel Core i7-9750H (6 physical cores, 12MB Cache, running up to 4.5 GHz) and 16GB RAM. To perform DNN training, the laptop is equipped with an NVIDIA GeForce RTX 2070 GPU with Max-Q Design (8GB RAM). The agents are evaluated with 11 trials, which are sufficient to detect meaningful differences between the methods while staying within our time and resource constraints. This adds confidence to the findings by showing consistent trends across these trials. The consistency of the results can be shown using box-and-whisker plots; they are graphical representations used to display data distribution and a summary of its key statistical properties. It can be used to show the minimums and maximums, median, quartiles, and potential outliers of multiple trials. For more details about the implementation of the simulated Fog environment and the DDQL hyper-parameters the reader is referred to \cite{EBRAHIM2023resilience, ebrahim2023privacyaware}.

\subsection{Results Discussion}
\label{subsec:results}
\begin{figure}[htbp]
\centering
\includegraphics[width=0.48\textwidth]{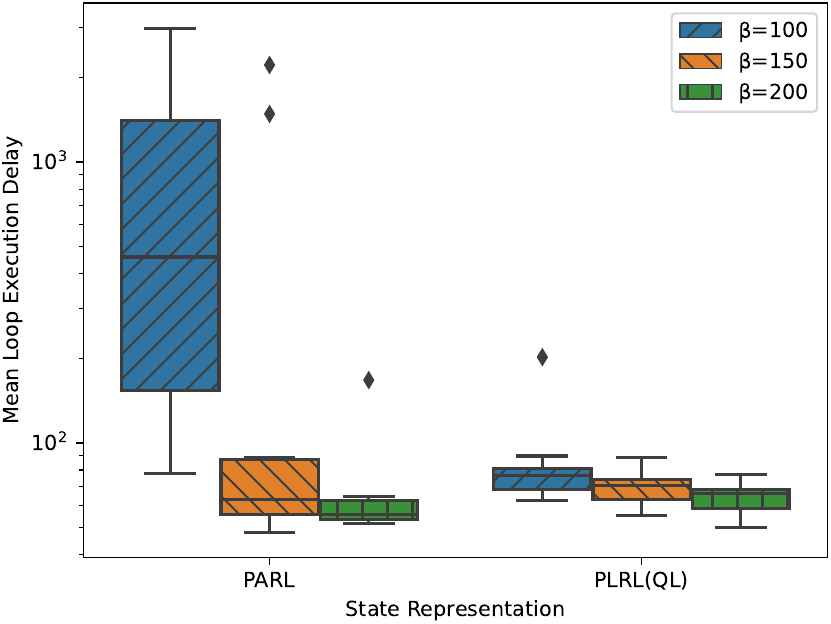}
\caption{Partially observable PARL vs. fully observable PLRL agents; trained with 30K training steps only.}
\label{fig:observability}
\end{figure}

Higher generation rates make it harder for PARL agents to reach the optimal solution \cite{ebrahim2023privacyaware}. This is because of the partial observability to maintain privacy, i.e., hiding Fog resource/load information. Fig. \ref{fig:observability} compares PARL and PLRL agents with 30K training steps to show how maintaining privacy can affect training time and convergence when trying to minimize execution delay. Unlike PARL, the PLRL agent that uses Queue Length (QL) only explicitly observes the number of waiting jobs on each Fog node as its state; it also receives the total number of waiting jobs in the whole system as a reward function for its actions \cite{ebrahim2023privacyaware}.

Fully observing the number of waiting jobs enables PLRL agents to train and converge faster because the agent can clearly see the impact of its last action in the environment. With low ($\beta=200$) and medium ($\beta=150$) generation rates, PARL agents are often able to converge with a short training time without the need to explicitly observe the exact number of waiting jobs in each Fog node. However, this partial observability with higher generation rates requires PARL agents to train longer to converge to the optimal solution. Since PARL agents easily converge with lower generation rates, we want to study the effect of TL from easier tasks, i.e., lower generation rates, to harder tasks, i.e., higher generation rates.

\begin{figure}[htbp]
\centering
\includegraphics[width=0.48\textwidth]{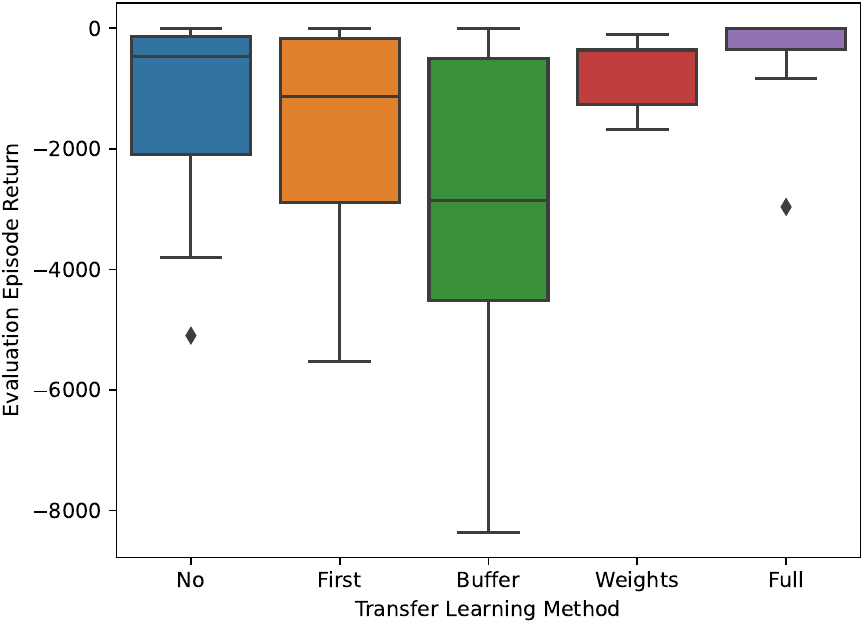}
\caption{Inference episode returns of PARL agents, evaluated with high rates for 100K simulation steps.}
\label{fig:eval}
\end{figure}

The use of TL in RL tasks with incremental levels of difficulty enables lifelong learning in environments that dynamically change over time. In addition, evaluating various TL techniques allows for a deeper understanding of what contributes to better performance and what contributes to negative guidance/transfer. Fig. \ref{fig:eval} shows the inference performance of each TL method on the final, i.e., hardest, task only. It shows the episode return for each method after trying to adapt to the three levels of difficulty during the lifetime of the agent.

The best performance is achieved using Full agent TL, where the minimum episode returns are achieved on almost half of the experiment trials (see Fig. \ref{fig:eval}). TL using just replay buffer experiences causes the agent to achieve the worst performance in most of the trials. This bad performance is an indication it is caused by biases that arise from the significant variations between previous experiences and the current condition of the environment. These biases are called negative guidance, which misguide the agent and hinder optimal learning in target tasks.

This performance degradation does not occur when only the knowledge of the agent (represented by the $Q$-network weights) is transferred. This is because the agent does not learn from irrelevant batches of experiences, avoiding the possibility of misguiding the agent in the new task. This TL technique achieves the second-best performance in our experiments. This can be explained by: 
\begin{itemize}
    \item \textbf{Source and Target Task Similarity:} If the agent has already learned valuable representations in the source task that are applicable to the target task, transferring $Q$-network weights can provide a head start in learning in the target task. Replay buffer experience TL, on the other hand, doesn't inherently capture the task similarity as explicitly.
    \item \textbf{Negative Transfer Risk:} Replay buffer experience TL can be prone to negative transfer, especially when source and target tasks are significantly different. Training on a diverse set of experiences might lead to the agent learning suboptimal policies or even counterproductive behavior in the target task. With weights TL, negatively guiding experiences are not there to misguide the agent.
    \item \textbf{Stability and Efficiency:} Transferring pre-trained $Q$-network weights provides a more stable starting point for learning in the target task. It can help the agent converge faster and require fewer samples to achieve good performance. Replay buffer experience TL might need more training iterations to achieve the same level of convergence since it starts with a randomly initialized network. This gets even worse if the replay buffer contains a mix of relevant and irrelevant experiences.
    \item \textbf{Domain Knowledge:} $Q$-network weights TL can leverage domain-specific knowledge effectively since a good behavior is already embedded in the agent. If there are certain features or patterns that are universally useful across tasks, transferring the weights helps retain and utilize this knowledge.
\end{itemize}

By comparing buffer TL against weights and Full agent TL, we conclude that fine-tuning previously learned weights using relevant or irrelevant interactions is much better than starting with randomized weights. However, we can still observe the effect of negative transfer in the marked outlier for Full agent TL (see Fig. \ref{fig:eval}); this is caused by previous irrelevant experiences. Since replay buffers are implemented with a finite capacity, new experiences will eventually replace older ones as the agent trains more in the target task. Therefore, more training eliminates the effect of negative guidance for Full agent TL, which allows it to consistently maintain superior performance compared to other techniques.

To underscore the impact of experience-induced negative guidance on PARL agents, we can contrast the performance of experience transfer with that of agents subjected to fresh retraining without TL. Retraining using randomized weights and newly gathered experiences following each incremental rate rise is superior to experience transfer. Avoiding negative guidance by retraining from scratch prevents potential learning interference and supports a more focused learning process tailored to the target task's requirements. However, it requires more training time to converge on the optimal solution since the agent starts without any pre-existing information, experiences, or biases. With more training, the agent adapts more specifically to the target task's dynamics, leading to gradual performance improvement over time. 

When it comes to improving the performance of an agent, it is more effective to adapt to environmental changes by retraining after each generation rate increase, whether from scratch or through TL. This fine-tuning approach is preferable to training the agent on the first task only, i.e., the easier task with the lowest generation rate, and then assessing its abilities on the more challenging task with higher generation rates. The agent, that adapts to environmental changes through lifelong fine-tuning, is able to handle the complexity of dynamic changes in demands; this makes the agent capable of delivering optimal results over the lifetime of the agent. However, transferring only buffer experiences can worsen the performance even more than agents that do not transfer anything, or even those that do not adapt, at all (see Fig. \ref{fig:eval}).

\begin{figure}[htbp]
\centering
\includegraphics[width=0.48\textwidth]{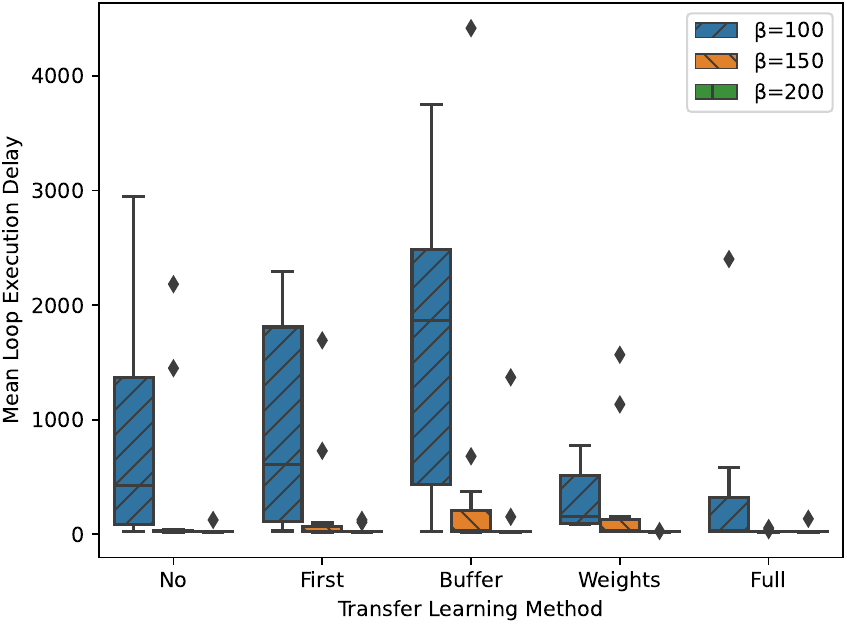}
\caption{PARL inference execution delay.}
\label{fig:delay}
\end{figure}

Fig. \ref{fig:delay} confirms the direct relationship between minimizing the waiting delay and minimizing the total system execution delay. It shows the mean system execution delay for each TL method. To examine the agent's performance, on each sub-task individually, we show the agent's inference performance before increasing the generation rate to the next difficulty level. The agent performs inference in a single episode of 100K simulation steps after being trained on each generation rate. The results in Figs. \ref{fig:delay} and \ref{fig:eval} are consistent, confirming our belief that minimizing the waiting delay is the key factor to minimizing the execution delay.

Fig. \ref{fig:delay} shows that Full and Buffer TL achieve the best and the worst execution delay, respectively. The bad performance that is achieved when only previous buffer experiences are transferred (Buffer TL) is caused by the biases of those experiences, which misguide the agent in the new environment. On the other hand, the second-best performance is achieved using Weights TL, i.e., when only the previous knowledge is transferred. This emphasizes the importance of knowledge transfer to achieve the best performance of Full TL, where previous domain-specific knowledge is effectively leveraged to mitigate the effect of negative guidance. 

Fig. \ref{fig:delay} also shows the inference performance on the medium and easiest tasks, i.e., with $\beta$ values of 150 and 200, respectively. This confirms that all methods can achieve better performance on easier tasks, i.e., with lower generation rates. Since learning is easier in easier tasks, it makes sense to leverage the knowledge learned in these tasks and embed it into the behavior of the agent while learning to adapt to the new environment. This shows the importance of TL in achieving efficient and effective adaptation to environmental change for lifelong learning agents. 

In summary, the findings suggest that fine-tuning previously learned weights through Weights TL or Full agent TL is more effective than starting with randomized weights. This helps improve agent performance and adaptability in a dynamic lifelong learning process. A more focused learning process can mitigate the remaining biases of irrelevant previous experiences in Full agent TL, which can be achieved with more training or smaller buffer size. In addition, these findings underscore the significance of minimizing waiting delay to reduce system execution delay in such a complex, heterogeneous, and unbalanced distributed computing environment.

\section{Conclusion}
\label{sec:conclusion}
Fog LB plays a crucial role in optimizing the performance, scalability, and fault tolerance of IoT applications. By intelligently distributing computational tasks across Fog nodes, LB algorithms ensure efficient resource utilization and low-latency communication. As the IoT landscape expands, Fog LB will remain a vital component in enabling real-time data processing, enhancing system efficiency, and delivering seamless user experiences in IoT environments. In this work, we explore different TL techniques to tackle such LB problems using lifelong learning RL agents. 

To the best of our knowledge, this is the first attempt to compare different TL approaches for the purpose of lifelong learning for Fog LB RL agents. TL is essential to efficiently adapt real-world RL solutions to environmental changes, including abrupt changes in generation rates. Instead of a constant rate, it is more realistic to introduce significant variations in generation rates over time, mimicking peak usage periods such as during the daytime, during special events, or because of an increase in the number of users and devices. In these situations, generation rates can get much higher, making the learning task much harder for Fog LB RL agents. 

The difficulty increases when considering the privacy required by Fog service providers to hide Fog load/resource information, which renders the environment partially observable. Privacy is one of the main challenges in real-world Fog LB solutions, which can be maintained through unique state/reward representations that do not require this information. In addition, adapting to significant environmental changes and minimizing the action delay of RL agents are two very important challenges to the real-world deployment of RL solutions. 

Using TL for PARL agents that are deployed in real-world environments allows them to deal with these challenges. Since training is very expensive for RL agents, it is often performed in resource-rich devices. To provide real-time LB decisions, lightweight inference models are exported from trained agents to be deployed even in resource-limited devices. However, they must be tuned in case of significant environmental changes, which are not avoidable throughout the lifetime of the agent. Adaptability by model tuning can be achieved through TL in a lifelong learning process of the agent, which is more practical than continuous training.

In future work, we will study the effect of the number of training steps and the buffer size on achieving more consistent performance using Full agent TL. Having consistent performance is vital to provide semi-deterministic outcomes for end users in real-world environments, which is often hard to achieve using machine learning approaches. In addition, we will evaluate selective weight transfer, where only few layers from the policy neural network is transferred to differential between high-level and low-level learned features. We will also evaluate various privacy-aware representations of the state and reward to improve the performance even further while still maintaining the privacy of Fog nodes. Finally, we can explore the practical implementation of our lifelong learning approach in real-world IoT applications, aiming to bridge the gap between theoretical advancements and practical deployment.

\bibliographystyle{IEEEtran}
\bibliography{main.bib}

\begin{IEEEbiography}[{\includegraphics[width=1in,height=1.25in,clip,keepaspectratio]{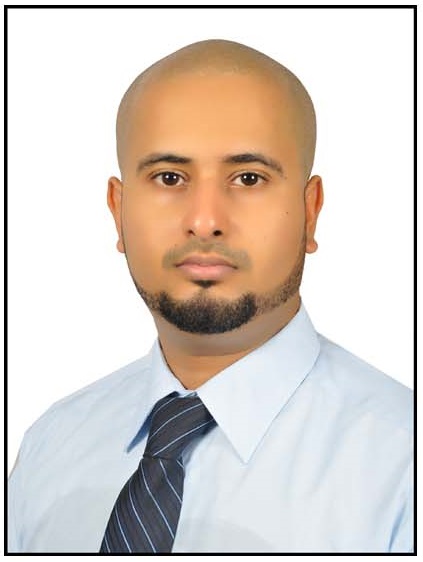}}]{Maad Ebrahim} is currently a Ph.D. candidate at the Department of Computer Science and Operations Research (DIRO), University of Montreal, Canada. He received his M.Sc. degree in 2019 from the Computer Science Department, Faculty of Computer and Information Technology, Jordan University of Science and Technology, Jordan. His B.Sc. degree in Computer Science and Engineering was received from the University of Aden, Yemen, in 2013. His research experience includes Computer Vision, Artificial Intelligence, Machine learning, Deep Learning, Data Mining, and Data Analysis. His current research interests include Fog and Edge Computing technologies, IoT, Reinforcement Learning, and Blockchains.
\end{IEEEbiography}

\begin{IEEEbiography}[{\includegraphics[width=1in,height=1.25in,clip,keepaspectratio]{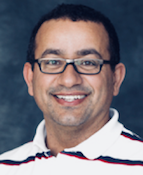}}]{Abdelhakim Senhaji Hafid}
spent several years as the Senior Research Scientist with Bell Communications Research (Bellcore), NJ, USA, working in the context of major research projects on the management of next generation networks. He was also an Assistant Professor with Western University (WU), Canada, the Research Director of Advance Communication Engineering Center (venture established by WU, Bell Canada, and Bay Networks), Canada, a Researcher with CRIM, Canada, the Visiting Scientist with GMD-Fokus, Germany, and a Visiting Professor with the University of Evry, France. He is currently a Full Professor with the University of Montreal. He is also the Founding Director of the Network Research Laboratory and Montreal Blockchain Laboratory. He is a Research Fellow with CIRRELT, Montreal, Canada. He has extensive academic and industrial research experience in the area of the management and design of next generation networks. His current research interests include the IoT, Fog/Edge Computing, blockchain, and intelligent transport systems.
\end{IEEEbiography}

\begin{IEEEbiography}[{\includegraphics[width=1in,height=1.25in,clip,keepaspectratio]{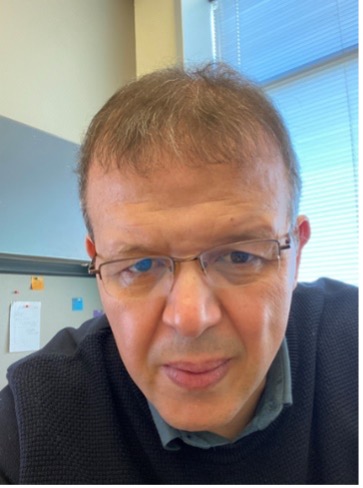}}]{Mohammed Riduan Abid}
earned his Ph.D. in computer science from Auburn University, USA, in 2010, as a Fulbright Student Scholar. He is now an Associate Professor at Columbus State University. Dr. Abid's accolades include a 2007 Fulbright Research Scholarship at the University of Houston and recognition as a USA National Academy of Sciences (NAS) Arab-American Fellow. In 2018, he joined Purdue University under an NAS Scholarship. With over 60 publications in prestigious conferences and journals, his research primarily focuses on cloud and edge computing, distributed systems, big data analytics, and high-performance computing (HPC). Beyond academics, Dr. Abid has a penchant for coaching and programming. His coaching prowess is evident from his six-year streak (2014–2019) of winning the National ACM Moroccan Collegiate Programming Contest (MCPC). Additionally, he led teams to the ACM International Collegiate Programming Contest (ICPC) World Finals twice, with appearances in Marrakesh in 2015 and Beijing in 2018.
\end{IEEEbiography}

\vfill
\end{document}